\documentclass[runningheads]{llncs}

\usepackage[mobile]{iciap}

\usepackage{iciapabbrv}

\usepackage{wrapfig}
\usepackage{graphicx}
\usepackage{booktabs}
\usepackage{placeins}
\usepackage[accsupp]{axessibility}  %

\usepackage{enumitem}
\usepackage{booktabs}   
\usepackage{multirow}   
\usepackage{makecell}
\usepackage{bm}
\usepackage{algorithm}
\usepackage[noend]{algpseudocode}
\algnewcommand\algorithmicinput{\textbf{Input:}}
\algnewcommand\algorithmicoutput{\textbf{Output:}}
\algnewcommand\Input{\item[\algorithmicinput]}
\algnewcommand\Output{\item[\algorithmicoutput]}

\usepackage{hyperref}

\usepackage{orcidlink}

\newcommand{\tit}[1]{\noindent\textbf{#1}~}

\begin{document}

\title{\texorpdfstring{How to Train Your\\Metamorphic Deep Neural Network}{How to Train Your Metamorphic Deep Neural Network}}

\author{Thomas Sommariva \and Simone Calderara \and Angelo Porrello}

\authorrunning{T. Sommariva et al.}

\institute{AImageLab, University of Modena and Reggio Emilia, Italy\\
\email{\{thomas.sommariva, simone.calderara, angelo.porrello\}@unimore.it}
}

\maketitle

\begin{abstract}
Neural Metamorphosis (NeuMeta) is a recent paradigm for generating neural networks of varying width and depth. Based on Implicit Neural Representation (INR), NeuMeta learns a continuous weight manifold, enabling the direct generation of compressed models, including those with configurations not seen during training. While promising, the original formulation of NeuMeta proves effective only for the final layers of the undelying model, limiting its broader applicability. In this work, we propose a training algorithm that extends the capabilities of NeuMeta to enable full-network metamorphosis with minimal accuracy degradation. Our approach follows a structured recipe comprising block-wise incremental training, INR initialization, and strategies for replacing batch normalization. The resulting metamorphic networks maintain competitive accuracy across a wide range of compression ratios, offering a scalable solution for adaptable and efficient deployment of deep models. 
\\ The code is available at: \url{https://github.com/TSommariva/HTTY_NeuMeta}
\end{abstract}

\section{Introduction and Related Works}
The increasing demand for deploying deep neural networks (DNNs) on edge devices or under real-time constraints underscores the challenge of dynamically adapting models to heterogeneous computational environments while maintaining high predictive accuracy. Addressing this issue is key to enabling broad deployment in real-world systems, including industrial systems and IoT devices, where balancing efficiency and performance is critical \cite{alajlan2022tinyml,capogrosso2024machine,matsubara2020head}.

To this end, several methods have been proposed to improve efficiency and adaptability. \textbf{Network Pruning}, reduces model size by removing parameters deemed less important, based on various importance heuristics \cite{Fang2023DepGraph,frankle2018lottery,hassibi1992second,he2018soft,li2016pruning,wen2016learning}. Another popular approach is \textbf{Knowledge Distillation}, where a compact student model is trained to replicate the behavior of a larger teacher model, thus maintaining accuracy while reducing inference cost \cite{buzzega2020dark,hinton2015distilling,romero2014fitnets,yuan2020revisiting,zhang2019your,zhang2018deep}. More recent research has focused on \textbf{Flexible Models}, which are trained to operate under multiple configurations, enabling dynamic adaptation to varying deployment scenarios. However, these methods are inherently limited to configurations observed during training, restricting their flexibility. \cite{cai2019once,chavan2022vision,grimaldi2022dynamic,hou2020dynabert,yu2018slimmable,yu2019universally}. 

More recently, \textbf{Neural Metamorphosis} (NeuMeta) \cite{yang2024neural} introduced a novel paradigm in which architectural hyperparameters -- such as the number of layers and channels -- can be adapted on the fly. Specifically, given a target architecture for deployment, the NeuMeta framework can dynamically generate the corresponding weights as a function of the selected hyperparameters. To do so, the authors of NeuMeta have proposed a \textbf{block-based} approach based on Implicit Neural Representation (INR) \cite{tancik2022blocknerf}. The overarching goal is to learn the \textbf{weight manifold} of a neural network, enabling the sampling of new networks of arbitrary sizes -- even for configurations not seen during training.

Despite its flexibility, the approach proposed by the authors of NeuMeta has been applied only to a limited portion of the network. For context, in ResNet-56 -- which contains approximately 27 residual blocks -- the method is used solely on the last. This highlights the complexity of extending metamorphosis to multiple layers, as also reflected in our experimental results. We observed that optimization becomes increasingly unstable as more blocks are incorporated into the morphing process. We attribute this trend to several potential shortcomings of NeuMeta, including its initialization strategy and the absence of batch normalization in the models sampled via INR.

In this work, we propose an improved training algorithm that extends the capabilities of NeuMeta \textbf{enabling full-network metamorphosis} of ResNet architectures \cite{he2016deep}. Our method adopts a gradual training strategy, applying neural metamorphosis to one residual block at a time. Building on this framework, we introduce several additional techniques to enhance training stability. For example, to accelerate convergence, each INR is initialized with the weights of the previously trained one. We also address the absence of batch normalization by incorporating learnable scaling coefficients, and further stabilize training across multiple architectures using gradient aggregation.

Experiments show that our approach significantly improves over the original NeuMeta framework, narrowing the accuracy gap with pre-trained models. We observe strong results on high-capacity networks, achieving near-baseline performance even under aggressive compression, an outcome of particular relevance from an efficiency-oriented deployment perspective. These findings establish the feasibility of full-network metamorphosis and lay the foundation for deploying dynamically adaptable DNNs across a wide range of real-world applications.

\section{Background}
In our work, we focus on ResNet56 \cite{he2016deep} and adopt the following nomenclature:
\begin{itemize}
\item \textbf{Block}: A residual block comprising two $3 \times 3$ convolutional layers, ReLU activations, and a shortcut connection.
\item \textbf{Layer}: A group of residual blocks that share the same number of output channels. Specifically, ResNet56 consists of three such layers of increasing width, with each layer that contains nine residual blocks.
\item \textbf{Base Layers}: Refers explicitly to parametric transformations, such as a convolutional layer or a linear layer.
\end{itemize}
\subsection{Neural Metamorphosis}
\label{subsec:Neural Metamorphosis}
NeuMeta \cite{yang2024neural} proposes a novel training framework that treats the set of weights of a neural network as points sampled from a continuous, high-dimensional weight manifold. NeuMeta aims to learn this manifold, enabling the direct generation of weights for neural networks with varying hyperparameter configurations (\eg, number of layers or channels).

To achieve this goal, NeuMeta employs a methodology based on Implicit Neural Representation (\textbf{INR}) \cite{reiser2021kilonerf}. An INR-based model is implemented through a Multi-Layer Perceptron (MLP), which takes as input a set of coordinates identifying a network configuration (see \cref{eq:Neumeta_input}) and outputs the corresponding weights. In particular, each INR receives as input a coordinate pair $\textbf{v} = (i, j)$, where $i = (L, C_{\text{in}}, C_{\text{out}})$ refers to a reference architecture at maximum capacity, and $j = (l, c_{\text{in}}, c_{\text{out}})$ corresponds to a sampled configuration with reduced capacity. In particular, $L$ denotes the number of base layers, while $l$ is the index of the base layer being predicted. The $l^\text{th}$ base layer of the reference architecture has $C_{in}$ input channels and $C_{out}$ output channels, while $c_{in}$ and $c_{out}$ are input$\backslash$output channels of the generated base layer.
Coordinates are normalized to capture relative proportions:
\begin{equation}
\label{eq:Neumeta_input}  
\textbf{v}=\left[\frac{l}{L}, \frac{c_{in}}{C_{in}}, \frac{c_{out}}{C_{out}},\frac{L}{N},\frac{C_{in}}{N},\frac{C_{out}}{N}\right] \text{, where }N=\underset{l\in L}{\max}(C_{in}^{(l)},C_{out}^{(l)}) .
\end{equation}
Coordinates are further enriched using Fourier positional embeddings to facilitate learning high-frequency details \cite{tancik2020fourier}. 

To sample the weights across all layers, NeuMeta employs a block-based approach \cite{tancik2022blocknerf}, where each weight and each bias of every base layer is modeled by a separate INR. All INRs share a common architecture, including output dimensionality $\mathcal{D}$; however, their outputs are interpreted according to their target.
\begin{itemize}
\item \textbf{Convolutional kernels:} For kernels of size $k$, when $\mathcal{D} \geq k \times k$, the sampled kernel is obtained by extracting the central $k \times k$ window from the output.
\item \textbf{Bias:} The first value of the output vector is used as the bias.
\item \textbf{Linear layers:} Each projection weight is computed via a separate forward pass, with the final value given by the mean of the $\mathcal{D}$ predicted values.
\end{itemize}

\begin{figure}[t]
  \includegraphics[width=\textwidth]
  {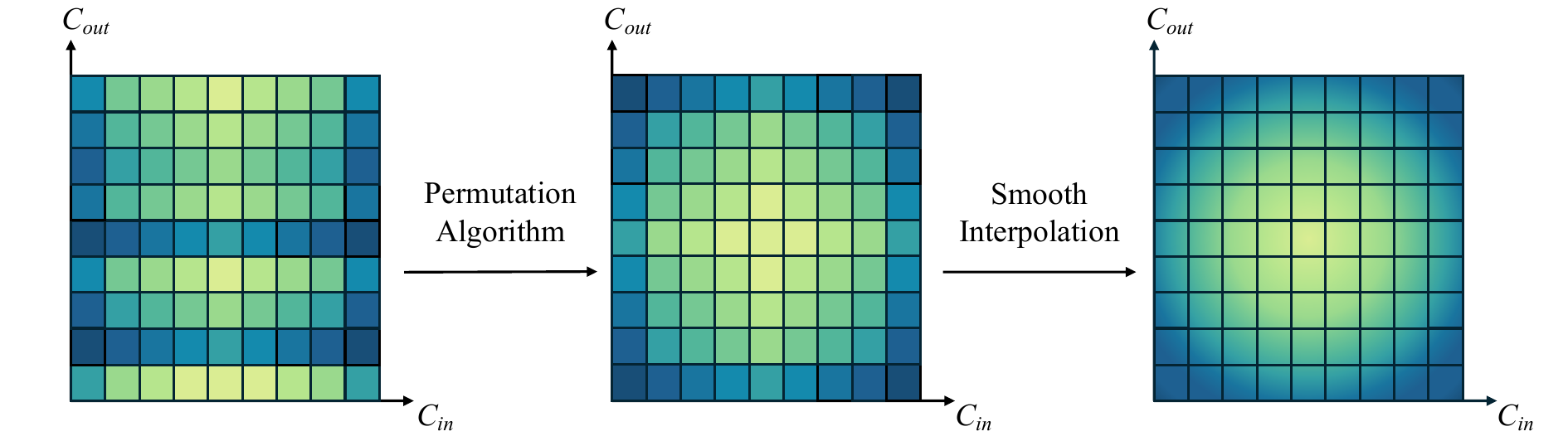}
  \caption{2D visualization of weight manifold smoothing via permutation alignment}
  \label{fig:permutation}
\end{figure}

\medskip
\tit{Smoothness of the weight manifold.} A primary challenge arises because INRs excel at learning low-frequency functions \cite{rahaman2019spectral,tancik2020fourier}, whereas the weight manifold of neural networks exhibits high-frequency variations. To address this discrepancy, NeuMeta leverages a pre-trained base model -- referred to as the \textbf{prior model} -- as a reference for supervising the training of the INR models. Smoothness is enforced through two \textbf{smoothing strategies} applied to the weights of the prior model: one promotes \textit{intra-network smoothness} (as illustrated in \cref{fig:permutation}), while the other encourages \textit{cross-network smoothness}. These strategies help the INR models better interpolate between neighboring configurations.
\begin{itemize}
    \item \textbf{Intra-network smoothing.} Consider a weight matrix $W \in \mathbb{R}^{C_{out}\times C_{in}}$, its local smoothness can be quantified using total variation (TV) \cite{yang2024neural}, defined as the variation along both channels $TV(W)=TV(W_{in}) + TV(W_{out})$. The authors of NeuMeta seeks a permutation matrix $P$ that minimizes the total variation of the resulting permuted weight matrix. Notably, the behavior of the corresponding base layer remains unaffected by such a permutation, provided that the permutation is inverted on the layer's output.
    \item \textbf{Cross-network smoothing.} To ensure that small differences in input configurations do not significantly affect the predicted parameters, a small perturbation $\epsilon \sim \mathcal{U}(-0.5, 0.5)$ is added to each input coordinate. This encourages the model to learn a smoother mapping by operating over a continuous space. Since the introduction of stochasticity renders the forward process non-deterministic, the model is evaluated multiple times with independently sampled perturbations. The final weights are then obtained by averaging the resulting predictions.
\end{itemize}

\tit{Batch normalization layers (BN).} During the metamorphic process, the BN layers of the prior model \cite{ioffe2015batch} are handled using a re-parameterization strategy \cite{ding2021repvgg}, which absorbs their operations—and thus their parameters—into adjacent base layers. This design is motivated by the fact that BN parameters capture activation statistics (mean and variance) that are inherently tied to the network architecture. As a result, statistics computed for one configuration may not be valid under different compression ratios (see \cref{eq:Gamma}). Re-parameterizing BN into neighboring base layers offers an architecture-agnostic alternative, ensuring compatibility across continuously changing network structures.
\subsection{Training Procedure of NeuMeta}
The training process of NeuMeta involves subsequent iterations of gradient-based optimization up to convergence. At each iteration, a configuration is sampled from a configuration pool, which is generated by applying varying compression ratios $\gamma\in[0.0,0.5]$ to the original architecture. The rate $\gamma$ is defined as:
\begin{equation}
\label{eq:Gamma} 
\gamma = 1 - \frac{\text{sampled channel number}}{\text{full channel number}}. 
\end{equation}
For each sampled configuration, the resulting weights are optimized with a composite loss function integrating three terms: a \textbf{task-specific loss}, a \textbf{reconstruction loss} and a \textbf{regularization loss}. Each term is weighted by a coefficient:
\begin{equation}
\label{eq:Neumeta_Loss} 
\mathcal{L}=\lambda_1\mathcal{L}_{\operatorname{TASK}}+\lambda_2\mathcal{L}_{\operatorname{RECON}}+\lambda_3\mathcal{L}_{\operatorname{REG}}.
\end{equation}
The \textbf{task-specific loss} $\mathcal{L}_{\operatorname{TASK}}$ evaluates the performance of the network generated by the INR models on the downstream task: for classification, it corresponds to the standard cross-entropy loss $\mathcal{L}_{\operatorname{TASK}} = - \sum_{k=1}^{K} y_k \log \hat{y}_k$. The \textbf{reconstruction term} $\mathcal{L}_{\operatorname{RECON}} = \| \theta_{\operatorname{PRIOR}} - \hat{\theta} \|_2^2$ encourages the predicted parameters $\hat{\theta}$ to stay close to the original, ideal weights $\theta_{\operatorname{PRIOR}}$ of the smoothed pre-trained model. This term is optimized only when the sampled configuration matches the prior model, \ie, when $\gamma = 0.0$. Finally, the \textbf{regularization component} $\mathcal{L}_{REG}=\|\hat{\theta} \|_2$ penalizes large weight magnitudes to mitigate overfitting \cite{loshchilov2017decoupled}.

\section{Method}
Empirically, we observe that NeuMeta struggles to learn the weight distribution across the entire network (see \cref{fig:fullNet_histo}). We attribute this limitation mainly to two factors: the challenge of learning the full weight manifold across multiple layers simultaneously, and the absence of batch normalization layers in the sampled models. To address these issues and further improve training effectiveness, we introduce a multi-step algorithm. Our training process aims to progressively metamorphose the entire architecture while preserving the performance of the original network. Each component of this method is detailed in the following subsections, while the full architectural design of the metamorphic block and a complete pseudocode are provided in the supplementary material.

\subsection{Incremental Training}
\label{subsec:IncrementalTraining}
As shown in our experiments (\cref{tab:LastBlock}), NeuMeta demonstrates effective metamorphosis of single residual blocks with minimal accuracy degradation. However, NeuMeta struggles when extending this approach to more layers. Inspired by this finding, we adopt a \textit{divide et impera} strategy, metamorphosing the network in a \textbf{block-wise, gradual} manner. 
Each stage of the training process focuses on transforming a single block into its metamorphic counterpart. During this phase, the previously learned INR models are jointly fine-tuned along with the current one. Moreover, the last block and the classification layer are not metamorphosed directly; instead, they are optimized across all sampled configurations to ensure consistent behavior of the last base layers under varying compression ratios.

This incremental approach allows each stage to focus on \textbf{localized} regions of the weight space, enabling more stable optimization and facilitating the learning of a continuous weight manifold across the architecture.
\subsection{Substitution of Batch Normalization Layers}
\label{subsec:SubstituteBN}
As discussed in \cref{subsec:Neural Metamorphosis}, NeuMeta removes \textbf{Batch Normalization} layers from the prior network. While removing BN has minimal impact when only a few blocks are transformed, applying it to larger portions of the network results in significant performance degradation—highlighting the critical role of batch normalization in training deep residual networks \cite{bjorck2018understanding,ioffe2015batch}.

To address this issue, we mitigate the absence of normalization layers by modifying the residual paths, drawing inspiration from \cite{de2020batch}. These modified blocks preserve the standard residual structure but incorporate a \textbf{learnable coefficient} $\boldsymbol{\alpha}$, initialized to zero, which scales the residual branch before it is added to the skip connection. This is motivated by the observation that BN layers typically downscale residual activations during early stages of training. Consequently, the main contribution to the propagated signal is provided by the skip connections, causing residual blocks to approximate identity mappings. Hence, introducing $\alpha$ emulates the activations of BN-equipped networks, ensuring stable training without explicit normalization.
\subsection{INR Pre-Initialization}
\label{subsec:CustomInit}
While the original NeuMeta's framework initializes each INR model randomly, our incremental training paradigm enables a more effective strategy. Specifically, we initialize each newly added INR model with the final weights of its predecessor. This approach enables knowledge transfer across INR networks trained subsequently, leading to faster convergence, as each block can build upon the representations learned in earlier blocks.
\subsection{Gradient Accumulation}
\label{subsec:GradientAccumulation}
The NeuMeta training procedure updates the weights of the INR based on gradients computed from a single sampled configuration. We argue that this strategy could bias the current optimization step toward the sampled configuration, limiting the generalization ability of the INR model. To mitigate this, we adopt gradient accumulation, aggregating losses over \textbf{multiple configurations} before performing a weight update. This strategy improves gradient quality and supports more effective joint learning across architectural variants.

Additionally, each mini-batch includes the uncompressed configuration ($\gamma=0.0$), ensuring that the reconstruction term in the loss function (see  \cref{eq:Neumeta_Loss}) contributes to every backward pass.
\subsection{Disentangling Weights and Biases}
\label{subsec:DisentanglementWB}
As discussed in \cref{subsec:Neural Metamorphosis}, the original block-based INR approach differentiates between weights and biases only after inference through selective extraction of the relevant output portion. We argue that this design introduces representational ambiguity, especially when the hypernetwork generates outputs that are mostly discarded, as happens in bias predictions.

To address this issue, we propose disentangling weights and biases at the architectural level by employing separate INRs for each. In this respect, every INR is designed with an output dimensionality tailored to its target: scalars for biases and full $3 \times 3$ matrices for convolutional kernels.

\section{Experiments}
\subsection{Implementation Details}
We use ResNet-56 \cite{he2016deep} as the backbone architecture, trained on the CIFAR100 dataset \cite{krizhevsky2009learning}, which provides sufficient complexity to validate the effectiveness of our proposed method. During training, input images are augmented via random cropping and horizontal flipping.
Two variants of the ResNet-56 architecture are evaluated: the first configuration (\cref{tab:Main} left side) is a lightweight model with 16, 32, and 64 channels in the first, second, and third layers, respectively; the second variant (\cref{tab:Main} right side) features a higher-capacity design with 64, 128, and 256 channels in the corresponding layers. For computational efficiency, we apply the metamorphosis only to the final layer of the architecture (for a total of \textbf{7 blocks}). This represents an extension over NeuMeta, which modifies only \textbf{the last block}; however, our method can be easily extended to earlier layers with minimal accuracy degradation (see \cref{fig:fullNet_histo,fig:accVSBlocks}).

The base model of each INR network consists of an eight-layer MLP with ELU activation functions \cite{clevert2015fast} and residual connections, each layer contains 512 neurons, and the positional embedding frequency is set to 32. Optimization is performed using AdamW \cite{loshchilov2017decoupled}, a warm-up phase of 20 epochs linearly increases the learning rate from $0$ to $8 \times 10^{-4}$, after which it remains constant. Considering the coefficients employed in the loss function \cref{eq:Neumeta_Loss}, we set $\lambda_1=10^2$, $\lambda_2=1$, and $\lambda_3=10^{-3}$. Each stage of our incremental training strategy lasts for 50 epochs (\cref{subsec:IncrementalTraining}); in the non-incremental settings, the model is trained for the same total number of iterations as the baseline method. Gradient Accumulation (\cref{subsec:GradientAccumulation}) is performed on a mini-batch of 4 configurations. We assign to each block a unique $\alpha$, shared across all compression ratios.

Experiments are primarily conducted on a single Nvidia RTX 6000 GPU with 24 GB of VRAM. For the higher-capacity ResNet configuration, due to increased memory requirements for storing all model configurations at that scale, we utilize an Nvidia A40 GPU with 48 GB of VRAM.
\subsection{Baselines}
We compare our approach against the following methods:
\begin{itemize}
    \item \textbf{Neural Metamorphosis} \cite{yang2024neural}: Serves as our primary baseline. Since our framework builds upon NeuMeta, this comparison is equivalent to an ablation of all the new components introduced in this work.
    \item \textbf{Individually Trained ResNet:} Separate ResNet models are trained from scratch at each compression ratio -- denoted as \textit{Individual} in our experiments.
    \item \textbf{Pruning:} We evaluate structural pruning techniques, including random pruning, Hessian-based pruning, Taylor-based pruning, and magnitude-based pruning ($l_1$ and $l_2$ norms). Pruned networks are obtained using the Dependency Graph method \cite{Fang2023DepGraph}, which constructs the inter-layer dependency graph to enable group-wise pruning of structurally coupled parameters.
\end{itemize}
\subsection{Results and Ablative Studies}
\begin{table}[t]
\caption{Ablation study across different $\gamma$. $^\dagger \gamma=0.75$ wasn’t applied in training.}
\label{tab:Main}
\setlength\tabcolsep{3pt}
\scriptsize
\centering
\begin{tabular}{l|ccccc|ccccc}
\midrule
& \multicolumn{5}{c|}{\textbf{ResNet56}} 
& \multicolumn{5}{c}{\textbf{ResNet56 with higher capacity}} \\
\cline{2-11}
& \makecell{$\gamma:$\\ 0.0} & \makecell{$\gamma:$\\0.25} & \makecell{$\gamma:$\\0.5} & \makecell{$\gamma:$\\$0.75^\dagger$ }
& \makecell{Train\\Cost}
& \makecell{$\gamma:$\\ 0.0} & \makecell{$\gamma:$\\0.25} & \makecell{$\gamma:$\\0.5} & \makecell{$\gamma:$\\$0.75^\dagger$ }
& \makecell{Train\\Cost}\\
&Acc&Acc&Acc&Acc&GPUh
&Acc&Acc&Acc&Acc&GPUh \\ 
\midrule
Individual & 72.63& 71.50& 71.78& 70.59& 3.5 &74.86 & 74.73& 74.67& 74.82& 5.1\\
NeuMeta \cite{yang2024neural} & 49.03&48.69& 48.65&45.38& 19.8& 69.07& 69.01& 68.90& 68.84& 57.1 \\
\rowcolor{gray!21} Ours & \textbf{66.13}&\textbf{ 66.00}& \textbf{66.10}& \textbf{65.36}& 16.2&\textbf{71.69}& \textbf{71.66}& \textbf{71.60}& \textbf{71.67}& 46.8\\
$-$ Incremental training&57.16 &57.24 &56.92 &53.22 & 19.6&70.61 & 70.72& 70.59& 70.70& 56.7 \\
$-$ $\alpha$ scaling&62.56 &62.68 &62.56 & 61.35& \textbf{16.1}& 71.02& 70.93& 71.03& 71.01& \textbf{46.6} \\
$-$ Grad Accumulation & 64.54 &64.71 &64.48 &63.78 & 17.7&71.15 &71.16 &71.14 &71.15 &53.2\\
$-$ Initialization&64.81 &64.76 &64.67 &63.37 & 16.2&71.04 &71.04 &71.09 &71.01 & 46.8\\
$-$ DisentaglmentW\&B& 65.45& 65.26&65.24 &64.74 & 16.3 & 71.09& 71.11& 71.10& 71.03& 46.9\\
\bottomrule
\end{tabular}

\resizebox{\textwidth}{!}{
}
\end{table}
\begin{figure}[t]
  \centering
  \begin{minipage}[t]{0.60\textwidth}
    \vspace{0pt}
    \scriptsize
    \setlength\tabcolsep{2pt}
    \captionof{table}{Ablation study on the last block of the third layer of ResNet56. $^\dagger \gamma=0.75$ wasn’t applied in training.}
    \label{tab:LastBlock}
    \centering
    \resizebox{\textwidth}{!}{
    \begin{tabular}{l|ccccc}
\midrule
& \multicolumn{5}{c}{\textbf{One metamorphic block}} \\
\cline{2-6}
& \makecell{$\gamma:$\\ 0.0} & \makecell{$\gamma:$\\0.25} & \makecell{$\gamma:$\\0.5} & \makecell{$\gamma:$\\$0.75^\dagger$}
&\makecell{Train\\Cost} \\
&Acc&Acc&Acc&Acc&GPUh \\ 
\midrule
Individual & 72.63 & 71.94& 72.31& 71.99&3.5 \\
NeuMeta \cite{yang2024neural} &71.84 &71.82 &71.79 &71.76&1.8\\
\rowcolor{gray!21} Ours &\textbf{72.24} &\textbf{72.23} &\textbf{72.24} &\textbf{72.23}&\textbf{1.4} \\
$-$ $\alpha$ scaling  & 72.17& 72.14& 72.14& 72.17&1.4\\
$-$ Grad Accumulation&72.22 &72.21 &72.21 &72.20&1.5\\
$-$ DisentaglmentW\&B&72.21 &72.19 & 72.20& 72.19&1.4\\
\bottomrule
\end{tabular}
 
    }
  \end{minipage}
  \hfill
  \begin{minipage}[t]{0.38\textwidth}
    \vspace{0pt}
    \centering
    \includegraphics[width=\linewidth]{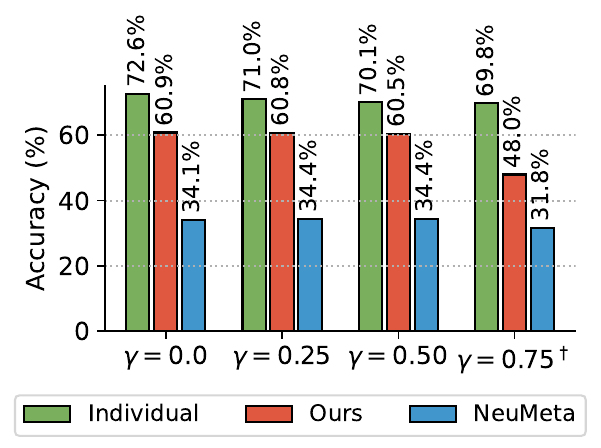}
    \caption{Full-network metamorphosis across different $\gamma$.}
    \label{fig:fullNet_histo}
\end{minipage}
\end{figure}

\begin{figure}[t]
\begin{minipage}[t]{0.48\textwidth}
    \centering
    \includegraphics[width=\textwidth]
  {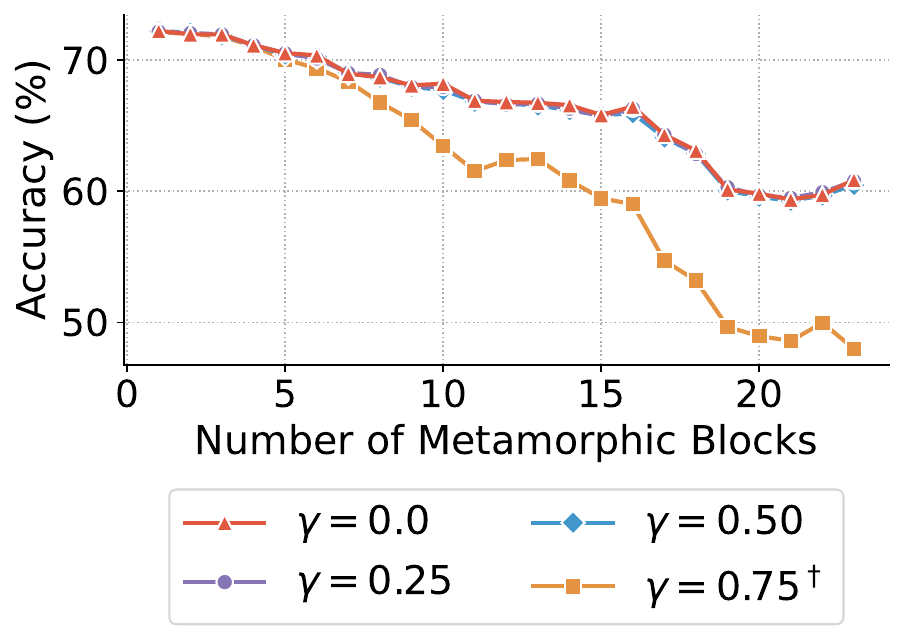}
    \caption{Accuracy across varying numbers of metamorphic blocks for different $\gamma$.}
    \label{fig:accVSBlocks}
\end{minipage}
\hfill
\begin{minipage}[t]{0.48\textwidth}
    \centering
    \includegraphics[width=\textwidth]{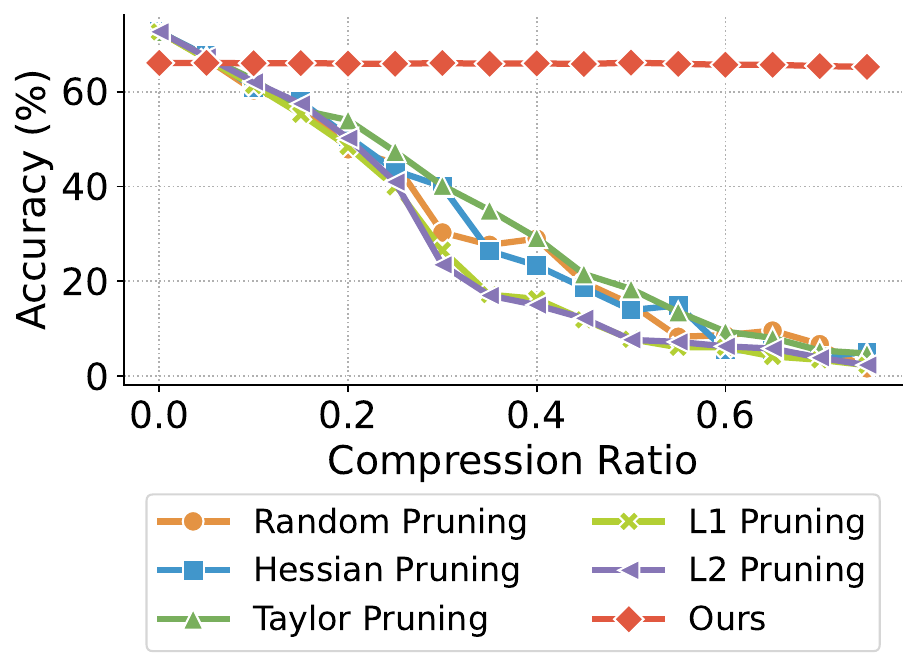}
    \caption{Accuracy comparison with structural pruning techniques.}
    \label{fig:Pruning}
  \end{minipage}
\end{figure}
\Cref{tab:Main} presents the top-1 accuracy and training time for our experiments applying metamorphosis to the final layer of ResNet-56. Additionally, \cref{tab:LastBlock} reports results for the setting where metamorphosis is restricted to the last block, consistent with the original setup proposed in \cite{yang2024neural}. Notably, our results indicate that the original NeuMeta approach struggles in extended scenarios as evidenced by its performance drop (see \cref{tab:Main} and \cref{fig:fullNet_histo}) -- highlighting the challenges of applying neural metamorphosis across multiple blocks. This effect is further illustrated in \cref{fig:accVSBlocks}, which depicts the progressive accuracy degradation as the number of metamorphic blocks increases.
With respect to the training times reported in \Cref{tab:Main,tab:LastBlock}, each line corresponds to the training time of the hypernetwork, which generates the weights for all $\gamma$. For the line “Individual”, instead, the reported time reflects the cumulative training time all ResNet models, each trained independently at a specific value of $\gamma$.

Although a performance gap remains with respect to individually trained models (see \textit{Individual}, first row of \cref{tab:Main,tab:LastBlock}), our training algorithm significantly outperforms NeuMeta \cite{yang2024neural}. Specifically, our approach \textbf{improves overall accuracy}, \textbf{reduces training time}, and \textbf{enhances generalization} to unseen configurations (\ie, $\gamma=0.75^\dagger$). 

The experiments on the right side of \cref{tab:Main} demonstrate that our method scales effectively with network width, achieving improved performance in higher-dimensional settings. On the wider, \textbf{higher-capacity} ResNet, our approach consistently delivers strong results, often approaching the performance of individually trained models. This suggests that optimizing larger networks is intrinsically easier -- a trend also observed in NeuMeta, which narrows the performance gap compared to individually trained baselines. Nevertheless, our method still outperforms NeuMeta, even in this more favorable setting, reaching results close to that of the prior model. From an efficiency perspective, this is particularly advantageous, as compressing larger models yields significant resource savings.

To further evaluate the feasibility of \textbf{full-network metamorphosis}, \cref{fig:fullNet_histo} reports the accuracy of generated models when metamorphosis is applied to all residual blocks of ResNet-56, excluding the first block of each layer (those with projection shortcuts). These results confirm that full-network metamorphosis is achievable with our approach, resulting in only minor accuracy degradation.

We also compare our method with several structural \textbf{pruning strategies}. As shown in \cref{fig:Pruning}, across comparable compression settings, our approach consistently outperforms standard pruning techniques in terms of top-1 accuracy.

\medskip
\tit{Impact of each component.} Furthermore, both \cref{tab:Main,tab:LastBlock} present a fine-grained ablation study quantifying the contribution of each component in our framework. The \textbf{incremental training} strategy (\cref{subsec:IncrementalTraining}) has the most significant impact, improving accuracy by 12\% and narrowing the gap between $\gamma=0.0$ and the unseen $\gamma=0.75^\dagger$ from 4\% to under 1\%. It also reduces training time by about 20\%, as fewer parameters are generated during early stages. These results underscore the difficulty of modeling the full weight distribution and highlight the effectiveness of a progressive approach focused on localized subspaces.

Moreover, replacing BN layers with $\boldsymbol{\alpha}$ \textbf{scaling} (\cref{subsec:SubstituteBN}) improves accuracy by 3.5\%, reducing the gap between $\gamma=0.0$ and $\gamma=0.75^\dagger$ by 0.5\%. \textbf{Gradient accumulation} across multiple architecture configurations (\cref{subsec:GradientAccumulation}) contributes a 1.5\% accuracy gain and shortens training time by nearly 10\% (due to the fewer performed optimization steps), indicating that simultaneous optimization across compression ratios promotes better generalization.

Considering our proposed \textbf{INR initialization} strategy (\cref{subsec:CustomInit}), it yields a 1.3\% accuracy improvement on seen $\gamma$, and a 2\% improvement on $\gamma=0.75^\dagger$. This finding suggests that sequential weight transfer facilitates convergence to more meaningful features improving generalization.

Finally, \textbf{disentangling} the weights and biases (\cref{subsec:DisentanglementWB}) yields a modest yet meaningful 0.6\% accuracy boost, showing that dedicating separate modules to biases and kernel matrices avoids interference and improves learning dynamics.

\subsection{Generalization and Fine-Tuning Analysis}
\begin{figure}[t]
\begin{minipage}{0.47\textwidth}
  \includegraphics[width=\textwidth]
  {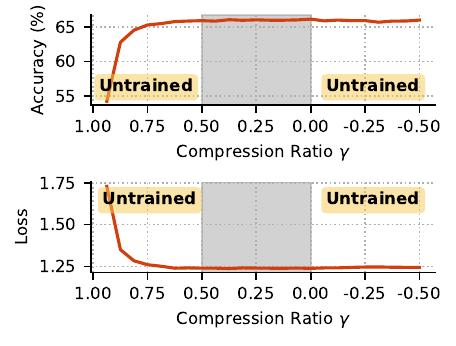}
  \vspace{-0.4cm}
  \caption{Accuracy and loss at different $\gamma$}
  \label{fig:accuracyVSdim}
\end{minipage}
\hfill
\begin{minipage}{0.49\textwidth}
  \includegraphics[width=\linewidth]{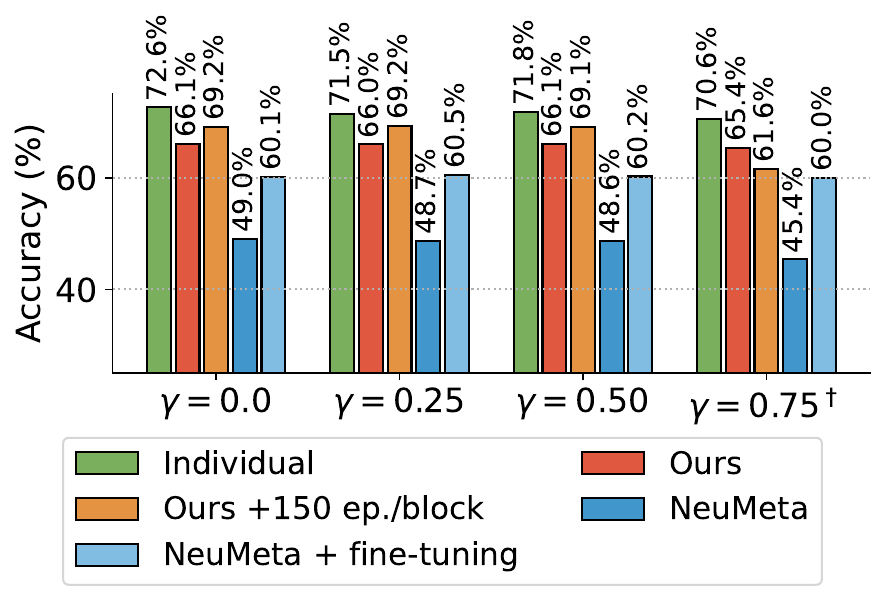}
  \caption{Accuracy comparison across different $\gamma$ for alternative training strategies.}
  \label{fig:histogram}
\end{minipage}
\end{figure}

In \cref{fig:accuracyVSdim}, we analyze performance trends across varying configurations. Notably, our approach maintains strong performance even for architectures with compression factors $\gamma$ smaller than those seen during training -- \ie, more powerful models. This clearly demonstrates that our approach can generalize to unseen architectures. Conversely, for smaller models, performance begins to degrade as compression rates approach or fall below $0.80$.

In \cref{fig:histogram}, we further investigate the ability of INR models on untrained configurations. Specifically, we examine the effect of extending the training duration by a factor of three. Interestingly, while prolonged training improves accuracy on configurations seen during training, it hampers generalization to unseen ones. This reveals a trade-off between in-distribution performance and flexibility across untrained $\gamma$ values. In contexts where adaptability is less critical, extended training can yield accuracy comparable to that of individually trained models.

Lastly, \cref{fig:histogram} also analyzes the effect of training the INR with the baseline method proposed in \cite{yang2024neural}, augmented with the fine-tuning strategy applied to the last block of ResNet and the final classifier, as described in \cref{subsec:IncrementalTraining}. The results demonstrate that the performance of the generated models can be significantly enhanced by simply adapting the final segment of the network to the preceding metamorphic blocks.

\section{Conclusion}
We focus on metamorphic neural networks -- a class of models that can adjust their weights to fit desired network configurations. We observe that their training stability deteriorates when multiple blocks are transformed. To address this, we propose a multi-stage training algorithm that progressively enables the generation of fully metamorphic networks. While a performance gap remains between our generated models and individually trained counterparts, we show that our approach can be further improved by employing high-capacity prior networks or by extending the training duration. Our findings demonstrate the feasibility of full-network metamorphosis in ResNet architectures, and we hope this work paves the way for future research into extending metamorphic capabilities to a broader range of architectures, like Recurrent Neural Networks~\cite{2018IPAS_QRNN,gers2000learning} and ViT.

\bibliographystyle{splncs04}
\bibliography{main}

\begin{thebibliography}{10}
\providecommand{\url}[1]{\texttt{#1}}
\providecommand{\urlprefix}{URL }
\providecommand{\doi}[1]{https://doi.org/#1}

\bibitem{alajlan2022tinyml}
Alajlan, N.N., Ibrahim, D.M.: Tinyml: Enabling of inference deep learning models on ultra-low-power iot edge devices for ai applications. Micromachines  \textbf{13}(6), ~851 (2022)

\bibitem{bjorck2018understanding}
Bjorck, N., Gomes, C.P., Selman, B., Weinberger, K.Q.: Understanding batch normalization. NeurIPS  \textbf{31} (2018)

\bibitem{2018IPAS_QRNN}
Bolelli, F., Baraldi, L., Grana, C.: {A Hierarchical Quasi-Recurrent approach to Video Captioning}. In: 2018 IEEE International Conference on Image Processing, Applications and Systems (IPAS). pp. 162--167. IEEE (Dec 2018). \doi{https://doi.org/10.1109/IPAS.2018.8708893}

\bibitem{buzzega2020dark}
Buzzega, P., Boschini, M., Porrello, A., Abati, D., Calderara, S.: Dark experience for general continual learning: a strong, simple baseline. NeurIPS  \textbf{33},  15920--15930 (2020)

\bibitem{cai2019once}
Cai, H., Gan, C., Wang, T., Zhang, Z., Han, S.: Once-for-all: Train one network and specialize it for efficient deployment. arXiv preprint arXiv:1908.09791  (2019)

\bibitem{capogrosso2024machine}
Capogrosso, L., Cunico, F., Cheng, D.S., Fummi, F., Cristani, M.: A machine learning-oriented survey on tiny machine learning. IEEE Access  \textbf{12},  23406--23426 (2024)

\bibitem{chavan2022vision}
Chavan, A., Shen, Z., Liu, Z., Liu, Z., Cheng, K.T., Xing, E.P.: Vision transformer slimming: Multi-dimension searching in continuous optimization space. In: ICCV. pp. 4931--4941 (2022)

\bibitem{clevert2015fast}
Clevert, D.A., Unterthiner, T., Hochreiter, S.: Fast and accurate deep network learning by exponential linear units (elus). arXiv preprint arXiv:1511.07289  (2015)

\bibitem{de2020batch}
De, S., Smith, S.: Batch normalization biases residual blocks towards the identity function in deep networks. NeurIPS  \textbf{33},  19964--19975 (2020)

\bibitem{ding2021repvgg}
Ding, X., Zhang, X., Ma, N., Han, J., Ding, G., Sun, J.: Repvgg: Making vgg-style convnets great again. In: CVPR (2021)

\bibitem{Fang2023DepGraph}
Fang, G., Ma, X., Song, M., Mi, M.B., Wang, X.: Depgraph: Towards any structural pruning. In: CVPR. pp. 16091--16101 (June 2023)

\bibitem{frankle2018lottery}
Frankle, J., Carbin, M.: The lottery ticket hypothesis: Finding sparse, trainable neural networks. arXiv preprint arXiv:1803.03635  (2018)

\bibitem{gers2000learning}
Gers, F.A., Schmidhuber, J., Cummins, F.: Learning to forget: Continual prediction with lstm. Neural computation  \textbf{12}(10),  2451--2471 (2000)

\bibitem{grimaldi2022dynamic}
Grimaldi, M., Mocerino, L., Cipolletta, A., Calimera, A.: Dynamic convnets on tiny devices via nested sparsity. IEEE Internet of Things Journal  \textbf{10},  5073--5082 (2022)

\bibitem{hassibi1992second}
Hassibi, B., Stork, D.: Second order derivatives for network pruning: Optimal brain surgeon. NeurIPS  \textbf{5} (1992)

\bibitem{he2016deep}
He, K., Zhang, X., Ren, S., Sun, J.: Deep residual learning for image recognition. In: CVPR (2016)

\bibitem{he2018soft}
He, Y., Kang, G., Dong, X., Fu, Y., Yang, Y.: Soft filter pruning for accelerating deep convolutional neural networks. arXiv preprint arXiv:1808.06866  (2018)

\bibitem{hinton2015distilling}
Hinton, G., Vinyals, O., Dean, J.: Distilling the knowledge in a neural network. arXiv preprint arXiv:1503.02531  (2015)

\bibitem{hou2020dynabert}
Hou, L., Huang, Z., Shang, L., Jiang, X., Chen, X., Liu, Q.: Dynabert: Dynamic bert with adaptive width and depth. NeurIPS  \textbf{33},  9782--9793 (2020)

\bibitem{ioffe2015batch}
Ioffe, S., Szegedy, C.: Batch normalization: Accelerating deep network training by reducing internal covariate shift. In: ICML (2015)

\bibitem{krizhevsky2009learning}
Krizhevsky, A., Hinton, G.: Learning multiple layers of features from tiny images. Tech. Rep.~0, University of Toronto, Toronto, Ontario (2009)

\bibitem{li2016pruning}
Li, H., Kadav, A., Durdanovic, I., Samet, H., Graf, H.P.: Pruning filters for efficient convnets. arXiv preprint arXiv:1608.08710  (2016)

\bibitem{loshchilov2017decoupled}
Loshchilov, I., Hutter, F.: Decoupled weight decay regularization. arXiv preprint arXiv:1711.05101  (2017)

\bibitem{matsubara2020head}
Matsubara, Y., Callegaro, D., Baidya, S., Levorato, M., Singh, S.: Head network distillation: Splitting distilled deep neural networks for resource-constrained edge computing systems. IEEE Access  \textbf{8},  212177--212193 (2020)

\bibitem{rahaman2019spectral}
Rahaman, N., Baratin, A., Arpit, D., Draxler, F., Lin, M., Hamprecht, F., Bengio, Y., Courville, A.: On the spectral bias of neural networks. In: ICML (2019)

\bibitem{reiser2021kilonerf}
Reiser, C., Peng, S., Liao, Y., Geiger, A.: Kilonerf: Speeding up neural radiance fields with thousands of tiny mlps. In: ICCV (2021)

\bibitem{romero2014fitnets}
Romero, A., Ballas, N., Kahou, S.E., Chassang, A., Gatta, C., Bengio, Y.: Fitnets: Hints for thin deep nets. arXiv preprint arXiv:1412.6550  (2014)

\bibitem{tancik2022blocknerf}
Tancik, M., Casser, V., Yan, X., Pradhan, S., Mildenhall, B., Srinivasan, P.P., Barron, J.T., Kretzschmar, H.: Block-nerf: Scalable large scene neural view synthesis. In: CVPR (2022)

\bibitem{tancik2020fourier}
Tancik, M., Srinivasan, P., Mildenhall, B., Fridovich-Keil, S., Raghavan, N., Singhal, U., Ramamoorthi, R., Barron, J., Ng, R.: Fourier features let networks learn high frequency functions in low dimensional domains. NeurIPS  \textbf{33} (2020)

\bibitem{wen2016learning}
Wen, W., Wu, C., Wang, Y., Chen, Y., Li, H.: Learning structured sparsity in deep neural networks. NeurIPS  \textbf{29} (2016)

\bibitem{yang2024neural}
Yang, X., Wang, X.: Neural metamorphosis. In: ECCV. Springer (2024)

\bibitem{yu2019universally}
Yu, J., Huang, T.S.: Universally slimmable networks and improved training techniques. In: iccv. pp. 1803--1811 (2019)

\bibitem{yu2018slimmable}
Yu, J., Yang, L., Xu, N., Yang, J., Huang, T.: Slimmable neural networks. arXiv preprint arXiv:1812.08928  (2018)

\bibitem{yuan2020revisiting}
Yuan, L., Tay, F.E., Li, G., Wang, T., Feng, J.: Revisiting knowledge distillation via label smoothing regularization. In: ICCV. pp. 3903--3911 (2020)

\bibitem{zhang2019your}
Zhang, L., Song, J., Gao, A., Chen, J., Bao, C., Ma, K.: Be your own teacher: Improve the performance of convolutional neural networks via self distillation. In: ICCV (2019)

\bibitem{zhang2018deep}
Zhang, Y., Xiang, T., Hospedales, T.M., Lu, H.: Deep mutual learning. In: CVPR (2018)

\end{thebibliography}

\clearpage
\appendix
\renewcommand{\thefigure}{A\arabic{figure}}
\setcounter{figure}{0}
\section*{Supplementary Material}
This supplementary material provides additional details for \textit{How to Train Your Metamorphic Deep Neural Network}. \cref{sec:metaBlock} describes the design of the metamorphic block, while \cref{sec:pseudocode} presents the complete pseudocode of our method.

\section{Design of the Metamorphic Block}
\label{sec:metaBlock}
\FloatBarrier
\begin{wrapfigure}{r}{0.4\textwidth}
  \centering
  \vspace{-20pt}
  \includegraphics[width=0.38\textwidth]{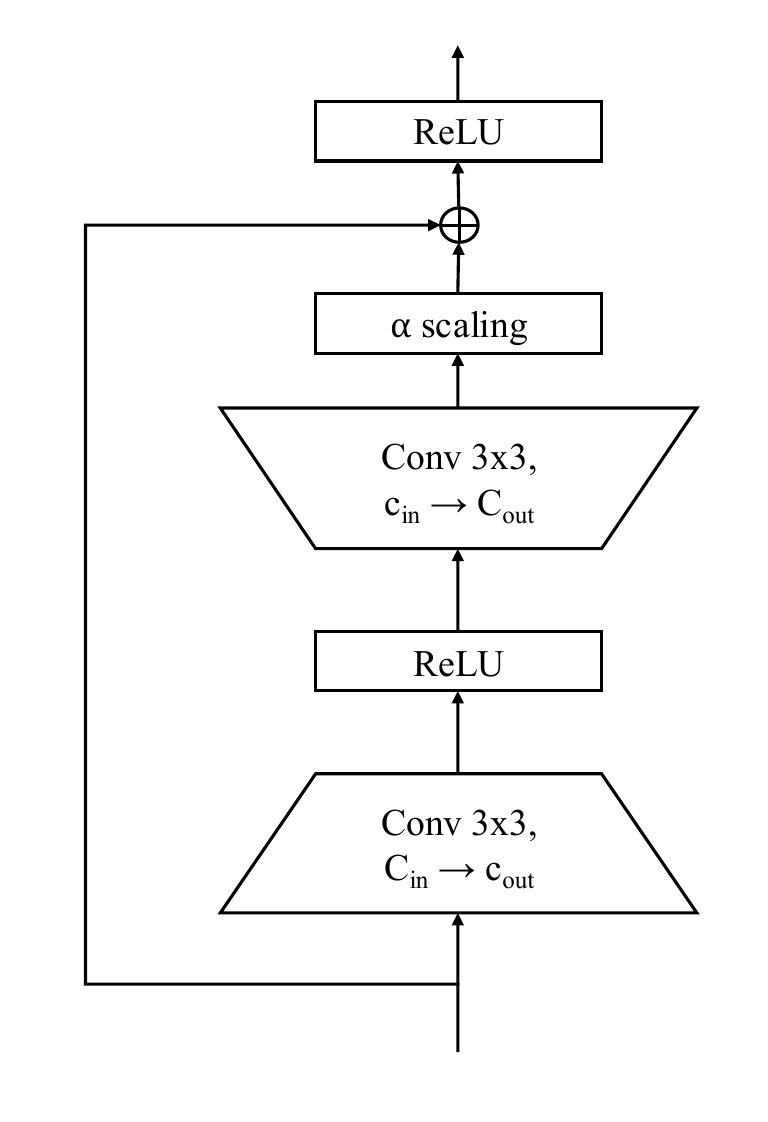}
  \caption{Design of the metamorphic block}
  \vspace{-20pt}
\end{wrapfigure}
In this section, we present the complete design of our metamorphic block, which integrates both the original structure proposed in NeuMeta \cite{yang2024neural} and the $\alpha$-scaling mechanism inspired by \cite{de2020batch}.

The metamorphic block consists of two $3\times3$ convolutional layers with a ReLU activation in between. The first convolution operates with $C_{in}$ input channels and $c_{out}$ output channels, where $C_{in}$ corresponds to the input dimensionality of the standard residual block \cite{he2016deep} in the same layer and $c_{out}$ is computed as $c_{out} = (1-\gamma)\times C_{out}$, effectively reducing the number of kernels proportionally to the compression ratio $\gamma$.

The second convolution takes $c_{in}$ input channels and outputs $C_{out}$ channels, restoring the original dimensionality. Consequently, the number of parameters in the second convolution is also reduced by a factor of $\gamma$, as it operates over a compressed intermediate feature space.
The output of the second convolution is scaled by the learnable coefficient $\alpha$ and added to the shortcut connection.

This design reduces the total number of parameters in the residual block by a factor of $\gamma$, through compression applied to both convolutional layers. Furthermore, the metamorphic block remains fully compatible with the standard residual structure, enabling the integration of standard and metamorphic blocks within the same network. This flexibility supports both partial metamorphosis and the incremental  strategy introduced in our training algorithm.

\clearpage
\subsection*{Training and Inference Pseudocode}
\label{sec:pseudocode}
\FloatBarrier
\begin{algorithm}[!h]
\caption{Neural Metamorphosis -- INR Training}
\label{algo:train}
\begin{algorithmic}[1]
\Input Smoothed ResNet: $f(\cdot;\theta_{\operatorname{PRIOR}})$, Training set: $D_{tr}=\{\mathbf{x}_i, y_i\}_{i=1}^M $, number of metamorphic blocks: $\#blocks$, accumulation steps: $\# acc$, INR lr: $\eta$, $\theta_{\operatorname{shared}}$ lr: $\eta_w$
\Output Implicit neural network $F(\cdot; W)$ and set of shared weights $\theta_{\operatorname{shared}}$
\For{$b$ \textbf{in} $\#blocks$}
\State $C_{out} \gets$ \# output channels of the b-th block of the prior model 
\State Define the configuration pool $\mathbf{I}\gets\{\frac{C_{out}}{2}, \frac{C_{out}}{2} + 1, \ldots, C_{out} \}$
\State Define the set of shared weights $\theta_{\operatorname{shared}}\gets\{\bm\alpha,\text{ classifier}\}$
\If{$b\neq0$}
\State Initialize $F(\cdot;W_{b})$ with $F(\cdot;W_{b-1})$
\EndIf
\For{$\operatorname{idx},(X,Y) \textbf{ in }enumerate(D_{tr}) $}
\State Sample a configuration $\mathbf{i} \in \mathbf{I}$ 

\For{$\mathbf{j}$ \textbf{such that} $\mathbf{j} \in \mathcal{J}_{\mathbf{i}}$} \Comment{$\mathcal{J}_{\mathbf{i}} := $ space of coordinates for \textbf{i}} 
\State $\mathbf{v} \gets \left[\frac{l}{L}, \frac{c_{\text{in}}}{C_{\text{in}}}, \frac{c_{\text{out}}}{C_{\text{out}}}, \frac{L}{N}, \frac{C_{\text{in}}}{N},  \frac{C_{\text{out}}}{N}\right] + \bm\epsilon$, where $\bm\epsilon\sim \text{U}(-0.5, 0.5)$

\State Generate weight : $\theta_{\mathbf{i},\mathbf{j}} \gets F(\mathbf{v};W)$
\EndFor

\State Inference with the generated weights: $\hat{Y} \gets f(X;\theta_{\mathcal{J}_{\mathbf{i}}})$
\State Compute total loss : $ \mathcal{L}\gets\lambda_1\mathcal{L}_{\operatorname{TASK}}+\lambda_2\mathcal{L}_{\operatorname{RECON}}+\lambda_3\mathcal{L}_{\operatorname{REG}} $
\State Update $\theta_{i}\text{ : }\theta_{i} \gets \theta_{i} - \eta_w \nabla_{\theta_{i}}\mathcal{L}$
\State Accumulate the scaled the loss : $\mathcal{L}_{\operatorname{AVG}} \gets \mathcal{L}_{\operatorname{AVG}} + \frac{\mathcal{L}}{s}$

\If{$\operatorname{idx} \% \ \# acc = 0$}
\State Compute INR gradients: $\nabla_{W} \mathcal{L}_{AVG} \gets \frac{\partial \mathcal{L}_{\operatorname{AVG}}}{\partial \theta_i} \frac{\partial \theta_i}{\partial W} $

\State Update the INR weights: $W \gets W - \eta \nabla_{W}\mathcal{L}_{\operatorname{AVG}}$
\EndIf
\EndFor
\EndFor
\end{algorithmic}
\end{algorithm}

\vspace{-25pt}
\begin{algorithm}[!h]
\caption{Neural Metamorphosis -- Weight Sampling}
\label{algo:inference}
\begin{algorithmic}[1]
\Input Trained INR $F(\cdot; W)$, Number of sampling iteration $K$, architecture configuration $\mathbf{i}$, set of shared weights $\theta_{\operatorname{shared}}$
\Output Weights $\theta_\mathbf{i}$ corresponding to configuration $\mathbf{i}$ 
\State Initialize $\theta_\mathbf{i}$ to $0$: $\theta_\mathbf{i} \gets \mathbf{0}$
\For{$k$ \textbf{in} $K$}
\For{$\mathbf{j}$ \textbf{such that} $\mathbf{j} \in \mathcal{J}_{\mathbf{i}}$} \Comment{$\mathcal{J}_{\mathbf{i}} := $ space of coordinates for \textbf{i}} 
\State $\mathbf{v} \gets \left[\frac{l}{L}, \frac{c_{\text{in}}}{C_{\text{in}}}, \frac{c_{\text{out}}}{C_{\text{out}}}, \frac{L}{N}, \frac{C_{\text{in}}}{N},  \frac{C_{\text{out}}}{N}\right] + \bm\epsilon$, where $\bm\epsilon\sim \text{U}(-0.5, 0.5)$

\State Generate and average weight : $\theta_{\mathbf{i},\mathbf{j}} \gets \theta_{\mathbf{i},\mathbf{j}} +\frac{1}{K} F(\mathbf{v};W)$
\EndFor
\EndFor
\State Load the shared weights in the generated network: $\theta_\mathbf{i} \gets \theta_{\operatorname{shared}}$
\end{algorithmic}
\end{algorithm}

\vspace{-10pt}
\FloatBarrier

\end{document}